\documentclass{article}
\usepackage[english]{babel}
\usepackage[letterpaper,top=2cm,bottom=2cm,left=3cm,right=3cm,marginparwidth=1.75cm]{geometry}
\usepackage{amsmath}
\usepackage{graphicx}
\usepackage[colorlinks=true, allcolors=blue]{hyperref}
\usepackage[utf8]{inputenc}
\usepackage[T1]{fontenc}
\usepackage{lineno}
\usepackage{soul}
\usepackage{todonotes}
\usepackage{nameref}
\usepackage{xcolor}
\usepackage{mwe}
\usepackage{graphicx,subcaption}
\usepackage{algorithm}
\usepackage{algpseudocode}
\usepackage[english]{babel}
\usepackage[strings]{underscore}
\usepackage{xcolor}
\usepackage{authblk}
\usepackage{cite} 
\usepackage{subcaption} 

\renewcommand\cite[1]{\textsuperscript{\!\citen{#1}}}

\title{\textbf{SCANIA Component X dataset: a real-world multivariate time series dataset for predictive maintenance}}

\author[1,*]{Zahra Kharazian}
\author[1,2]{Tony Lindgren}
\author[1]{Sindri Magnússon}
\author[2]{Olof Steinert}
\author[3]{Oskar Andersson Reyna}
\affil[1]{Stockholm University, Department of Computer and Systems Sciences, Kista, SE-164 07, Sweden}
\affil[2]{Scania CV, Strategic Product Planning and Advanced Analytics, Södertälje, SE-151 32, Sweden}
\affil[3]{Scania CV, Connected Intelligence, Södertälje, SE-151 32, Sweden}
\affil[*]{corresponding author: Zahra Kharazian (zahra.kharazian@dsv.su.se)}
\date{}

\begin{document}
\maketitle

\begin{abstract}

Predicting failures and maintenance time in predictive maintenance is challenging due to the scarcity of comprehensive real-world datasets, and among those available, few are of time series format. This paper introduces a real-world, multivariate time series dataset collected exclusively from a single anonymized engine component (Component X) across a fleet of SCANIA trucks. The dataset includes operational data, repair records, and specifications related to Component X, while maintaining confidentiality through anonymization. It is well-suited for a range of machine learning applications, including classification, regression, survival analysis, and anomaly detection, particularly in predictive maintenance scenarios. The dataset’s large population size, diverse features (in the form of histograms and numerical counters), and temporal information make it a unique resource in the field. The objective of releasing this dataset is to give a broad range of researchers the possibility of working with real-world data from an internationally well-known company and introduce a standard benchmark to the predictive maintenance field, fostering reproducible research.

\end{abstract}

\section{Background \& Summary}
In an era marked by technological advancement and data-driven decision-making, the automotive industry is undergoing a transformative shift, particularly in the realm of vehicle maintenance. Predictive maintenance (PdM) has emerged as a key player in revolutionizing how we care for and optimize the performance of vehicles. This innovative methodology harnesses the power of advanced analytics, sensor technology, and machine learning to predict when vehicle components are likely to fail, allowing for timely and cost-effective maintenance interventions.
PdM is paramount for critical components in trucks. This proactive approach to maintenance involves using data, analytics, and monitoring systems to estimate the components' health state. 

One of the significant challenges in PdM is the shortage of public real-world datasets. The reason is that Original Equipment Manufacturers (OEMs) tend to keep the data to themselves and do not share it with anyone outside their company. Exceptions are research or development partners who can access the data when working with companies for PdM solutions. Reasons for keeping the data hidden are that it is sensitive as it contains failure frequencies, what type of sensor is available, etc.
The lack of data makes researchers in the field use simulated datasets \cite{C-MAPSS,arias2021aircraft,voronov2018data,safdari2024synthetic} that mimic real conditions.
However, having access to real-world datasets remains essential for developing robust models for foreseeing real industrial equipment failures, as synthetic datasets usually lack the complexity of real-world data and overlook many common challenges, such as intricate relations between signals.

The newly released real-world dataset \cite{Lindgren2024SCANIA} from SCANIA is an exception to the typical practice of companies not releasing data publicly and holds significant potential for advancing the field of PdM.
Another unique advantage inherent in this dataset is its consideration of temporal information, demonstrating gradual degradation in equipment in the form of time series readouts.
This multi-variate time series dataset is likely rich in diverse information collected from SCANIA trucks and can be used for various tasks in PdM, such as classification, regression, forecasting, anomaly detection, and survival analysis.
The proposed dataset is introduced for the ongoing Industrial Challenge 2024 at the $22^{nd}$ International Symposium on Intelligent Data Analysis (IDA) with the title of \textit{Developing an Effective Predictive Model for Imminent Component X Failures in Heavy-Duty SCANIA Trucks} at Stockholm University, Sweden. The setting of this industrial challenge served as a nexus for collaboration, fostering a unique synergy between academia and industry. The proposed dataset opens the door to various opportunities for advancing research and optimizing PdM methodologies.

This dataset includes operational data, truck specifications, and the repair records of an anonymized truck engine component called component X. The component's name has been anonymized for proprietary reasons.
The wealth of information in the dataset empowers decision-makers to make informed choices regarding maintenance strategies, resource allocation, and fleet management. This dataset has been used in recent PdM-related research studies; for instance, Zhong et al. \cite{zhong2024implementing} implemented deep learning models such as convolutional and recurrent neural networks to detect failure patterns and maintenance cost management. Parton et al. \cite{parton2024predicting} employed a method using graph neural networks (GNNs) on the graphs that are derived from this time series data to estimate the remaining useful life (RUL).
Moreover, Carpentier et al. \cite{carpentier2024towards} applied different models in various applications like multiclass classification, regression, and survival analysis to predict component X's failure. Additionally, another study \cite{rahat2024survloss} used the dataset to evaluate a novel loss and error calculation method designed for survival analysis and RUL prediction. In a different application \cite{lindgren2022low}, it has been used for low dimensional synthetic data generation to enhance the predictive model's performance. Another study \cite{kharazian2024copal} highlights the effect of implementing active learning in enhancing the model's decision for RUL estimation on this dataset.

Another comparable dataset also comes from SCANIA, it was concerned with Air Pressure System (APS) \cite{miscapsfailureatscaniatrucks421} of trucks and was disclosed in a preceding industrial challenge 2016 at the 15th International Symposium on Intelligent Data Analysis (IDA), Stockholm University. The APS dataset is also suitable for many machine learning tasks in PdM field and has been used in various researches, such as \cite{huang2022energy, abidi2020automated, selvi2022air, lokesh2020truck, ranasinghe2019methodology, oh2024quantum, ke2022deep, sun2023robust, akarte2018predictive, rafsunjani2019empirical, syed2020novel,taghandiki2023minimizing,beikmohammadi2024Transformer}. 
Although the APS dataset is rich in the field, it does not capture the temporal information of variables. 

The rest of this paper is structured as follows: The Methodology section discusses the data collection process, associated challenges, and the steps taken to preserve data privacy through anonymization. The Data Records section provides a detailed description of the dataset. Finally, the Technical Validation section introduces a cost function designed to evaluate the performance of a trained model on the dataset.

\section{Methods}\label{sec:method}

\subsection{Data collection}\label{data collection}
The proposed dataset includes three sources of information 
encompassing important aspects of truck information. The first part comprises operational data. The second source contains repair records obtained from workshops. The last one incorporates the specifications of trucks.
Collecting such information from a fleet of trucks operating daily in different situations involves several challenges and potential errors.
Below, we explain how the data is collected and also mention some of the possible errors during data collection. Finally, we explain what considerations have been made to protect the confidentiality of data for publishing.

\subsubsection{Operational data}

For collecting operational data of this dataset, trucks' onboard sensors are utilized to monitor and collect crucial parameters like real-time data on the truck's condition and performance.
This data is stored inside the vehicle control units and is accessible remotely or using plugin cables in visiting workshops.
One challenge in compiling operational data is losing connection with devices that collect data. 
One example of this error happens with Electronic Control Units (ECUs), like the engine control unit; sometimes, when the ECU software is updated, the data collection counters could be reset, i.e., start again from the beginning. This type of problem with data is handled through post-processing of the data when it has been downloaded from the truck. The post-processing does cover the majority of the possible types of corrupt data when collecting information but it might not cover all cases.

\subsubsection{Repair records}
Repair records collected from trucks include information about maintenance, repairs, and servicing performed on the vehicles. This information is usually collected from invoices and work orders, including crucial information about services rendered, replaced parts, labor costs, and other expenses incurred during the workshop visits. If the components have been changed or marked as repaired, they will be considered as failed and labeled as such in the dataset. Otherwise, it will be considered as healthy. 
Here, the challenge is that SCANIA can receive information only from its own network and official workshops. 
Therefore, we limited the population of the vehicles to those with a complete service history.

\subsubsection{Specifications of trucks}
Specification data is collected with the production system. Where they provide detailed specifications for each truck's model the company has produced, this includes information about the engine type, weight capacities, dimensions, and other technical details.
One possible challenge in collecting such data is, in very rare cases, when the truck is rebuilt and no longer matches the original specification.

Some of the mentioned errors are handled or addressed using quality control measures and regular equipment maintenance. Also, collaboration between data scientists, engineers, and domain experts has been done as a crucial step for developing effective strategies to handle data collection challenges in PdM.

\subsection{Preserving Data Privacy} 

After collecting data, some modifications have been made to protect the confidentiality of data for publishing. Regarding temporal representation description, relative times are reported instead of the original timestamps. This is done to capture temporal patterns without revealing specific dates or time points. Repair frequencies and readout frequencies may have been modified and are not necessarily representative of actual truck usage.
Moreover, variable names have been omitted for privacy and proprietary reasons, presenting a subset of all available operational and specification data. Also, the dataset comprises a random subset of vehicles visiting SCANIA workshops, and real vehicle identity numbers are not disclosed and are reported as anonymous IDs. The chosen subset is representative of SCANIA workshop visits for Component X analysis.
In addition to these considerations, some perturbations like scaling have been made on operational data and repair rates. It is worth noting that the perturbed data still maintains its
utility for predictive modeling.        
To evaluate the robustness of the model, a sensitivity analysis is conducted, assessing how changes in perturbation levels impact the model's performance. This allowed us to observe model effectiveness variations under different perturbation conditions. Furthermore, it is validated that perturbations do not compromise the predictive model's accuracy.

The anonymization of the Component X dataset is necessary to protect proprietary and sensitive information, and without such modifications, it would not have been possible to publish this dataset.
While this process may add complexity to the interpretation of certain features, it is important to note that the anonymized data retains its utility for predictive modeling. Moreover, the dataset’s inherent temporal and operational patterns remain intact. These patterns further form a basis for training and evaluating predictive models. To facilitate the connection between anonymized data parts, a unique ID is associated with each vehicle across all shared files. This enables researchers to integrate information from different data parts, such as operational data and time-to-event information, ensuring that comprehensive analyses can still be conducted effectively despite anonymization. 

\section{Data Records}

The proposed dataset is publicly available and can be accessed on the 
Swedish National Data Service website
\href{ https://doi.org/10.5878/jvb5-d390}{(https://doi.org/10.5878/jvb5-d390)}\cite{Lindgren2024SCANIA}.
It is divided into three segments: training, validation, and testing. Each segment comprises multiple files in CSV (Comma-Separated Values) format, and this section provides detailed descriptions for each file to facilitate ease of use. 
Vehicles are randomly selected for inclusion in each segment to ensure the dataset's robustness and reliability, promoting a representative distribution across the entire dataset. This random selection process effectively allocates a specific percentage (70\%, 15\%, and 15\%) of vehicles to the training, validation, and testing sets, respectively, allowing for comprehensive model evaluation and development.

\subsection{Training set}
The training set contains three files including  "\textit{train\_operational\_readouts.csv}", "\textit{train\_tte.csv}", and "\textit{train\_specifications.csv}," where each is explained below.

\subsubsection{train\_operational\_readouts.csv} \label{sec:train op}

Operational data related to the train set is collected in a file named "\textit{train\_operational\_readouts.csv}". This data file comprises readouts collected at different times from a variety of features for a fleet of vehicles, yielding a multi-variate time series. 
These readouts, found in each row, offer unique insights into the operational status of these vehicles. Each readout encapsulates a distinct set of variables gathered between two points in time, such as $t_{i}$ and $t_{i+1}$.
In summary, it consists of 1122452 observations or instances from 23550 unique vehicles and 107 columns, including \textit{vehicle\_id} and \textit{time\_step}.
The column \textit{time\_step} acts as a gauge, measuring the duration in time\_step that each vehicle has been utilizing Component X during its operational lifespan. Note that vehicles do not necessarily follow the same sampling frequency in the time\_step column.
It is also worth noting that this dataset \cite{Lindgren2024SCANIA} does not encompass the entire available operational data but rather represents a carefully selected subset. Experts have specifically curated this subset, handpicking data they believe is most relevant. 
Ultimately, 14 attributes are selected and anonymized in the operational data, offering a broad spectrum of information without divulging specifics about the nature of Component X. 
These variables are organized into single numerical counters and histograms where each histogram has several bins and provides a compressed representation of the signal data by grouping the sensor readings into bins. The basis for the division of the bins varies from the range of the data (maximum and minimum of the variable), the variable characteristics, the resolution of the data, experts' knowledge, and industry-specific standards.
Each bin in a histogram represents certain conditions linked to the values observed within the measured features. 
For instance, imagine a histogram linked to the variable \textit{distance\_driven} with four bins representing \textit{ambient temperature}. This histogram shows the distribution of distance driven, organized into bins with different temperature ranges $\{(T< -20), (-20\leq T< 0), (0\leq T< 20), (T> 20)\}$. Each bar of this histogram shows a temperature range, and the height of each bar represents the frequency of \textit{distance\_driven} within that temperature range.
Histogram variables use the following indexing format: \textit{variableid\_binindex}. Where the "variableid" represents the ID of an anonymized variable or feature, and "binindex" shows the bin numbers.
As an example, the variable with "variableid" 167 is a multi-dimensional histogram that has ten bins, "167\_0", "167\_1",..., and "167\_9". 

\begin{figure}[h!]
    \centering
    \includegraphics[width=\linewidth]{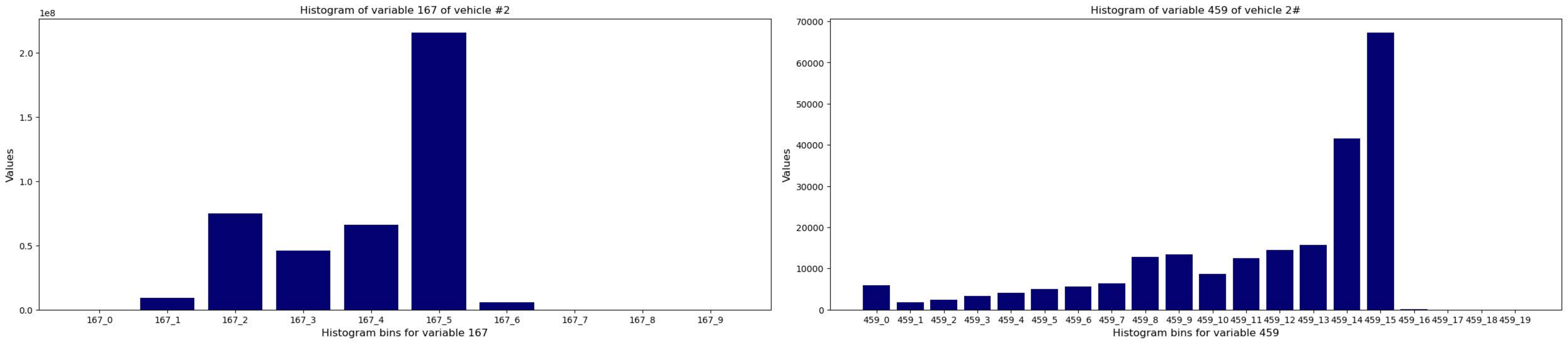}

    \begin{minipage}[t]{0.45\linewidth}
        \centering
        \vspace{-5mm}
        \subcaption{}\label{fig:1a}
    \end{minipage}
    \hfill
    \begin{minipage}[t]{0.45\linewidth}
        \centering
        \vspace{-5mm}
        \subcaption{}\label{fig:1b}
    \end{minipage}
    
    \caption{Visualization of two histogram variables for vehicle number 2. On the left, (a) displays histogram variable 167, while on the right, (b) illustrates histogram variable 459.}
    \label{fig:combined-image1}
\end{figure}

\begin{figure}[h!]
    \centering
    \includegraphics[width=\linewidth]{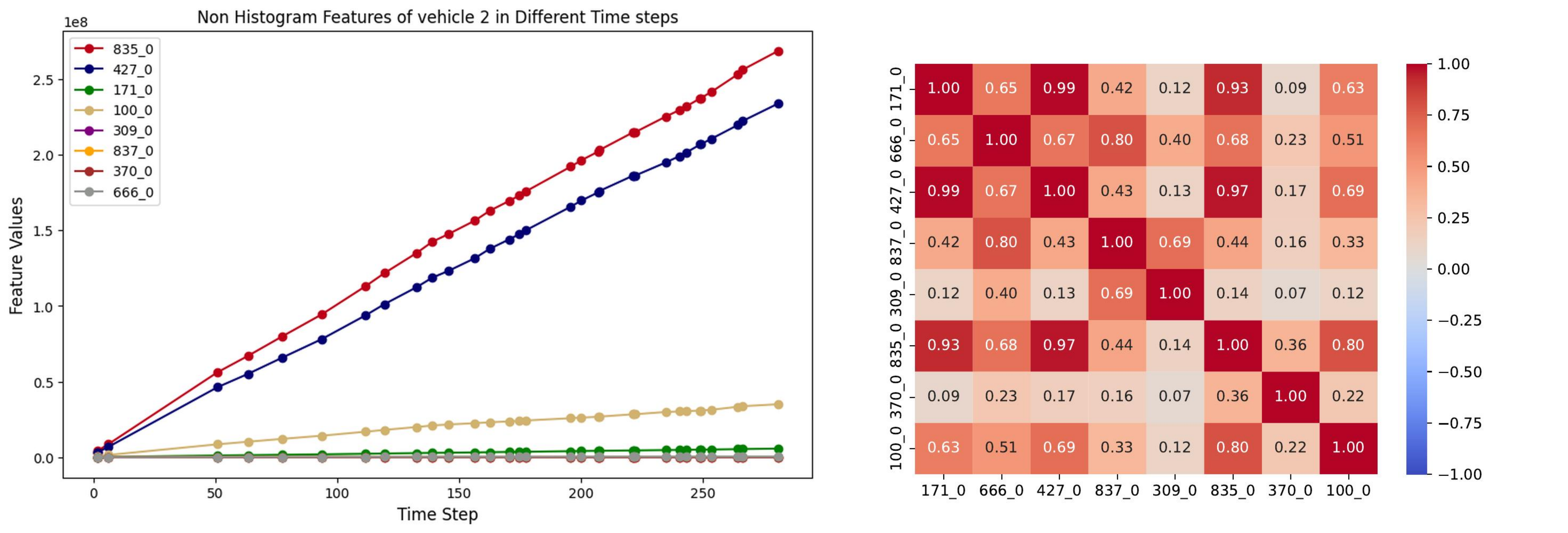}

    \begin{minipage}[t]{0.45\linewidth}
        \centering
        \vspace{-5mm}
        \subcaption{}\label{fig:2a}
    \end{minipage}
    \hfill
    \begin{minipage}[t]{0.45\linewidth}
        \centering
        \vspace{-5mm}
        \subcaption{}\label{fig:2b}
    \end{minipage}
    
    \caption{Analyzing non-histogram features over time for the train set. On the left, (a) demonstrates non-histogram variables for vehicle number 2 over time, while on the right, (b) shows the correlation of these features.}
    \label{fig:combined-image2}
\end{figure}

In summary, six out of 14 variables are organized into six histograms with variable IDs: "167", "272", "291", "158", "459", and "397," with 10, 10, 11, 10, 20, and 36 bins, respectively.
Figure \ref{fig:combined-image1} illustrates two histogram features of variables 
167 (see Fig. \ref{fig:1a}) and 459 (see Fig. \ref{fig:1b}) from an arbitrary vehicle at its last readout. 
Moreover, the eight rest of the variables named "171\_0", "666\_0", "427\_0", "837\_0", "309\_0", "835\_0", "370\_0", "100\_0"] are numerical counters. These features are mostly accumulative and are suitable for the representation of trends over time. Figure \ref{fig:combined-image2} visualizes these numerical counters in more detail. An example of these features for an arbitrary vehicle in this dataset \cite{Lindgren2024SCANIA} is depicted in Fig. \ref{fig:2a}.

Additionally, the correlation between these non-histogram features (numerical counters) is calculated considering all readouts of vehicles and illustrated in Fig. \ref{fig:2b}.  As can be seen, all the features are positively correlated, and no negative correlation exists between them. i.e., if the value of one feature increases, the value of the other feature also tends to increase.

\begin{figure}[h!]
\centering
\includegraphics[width=0.9\linewidth]{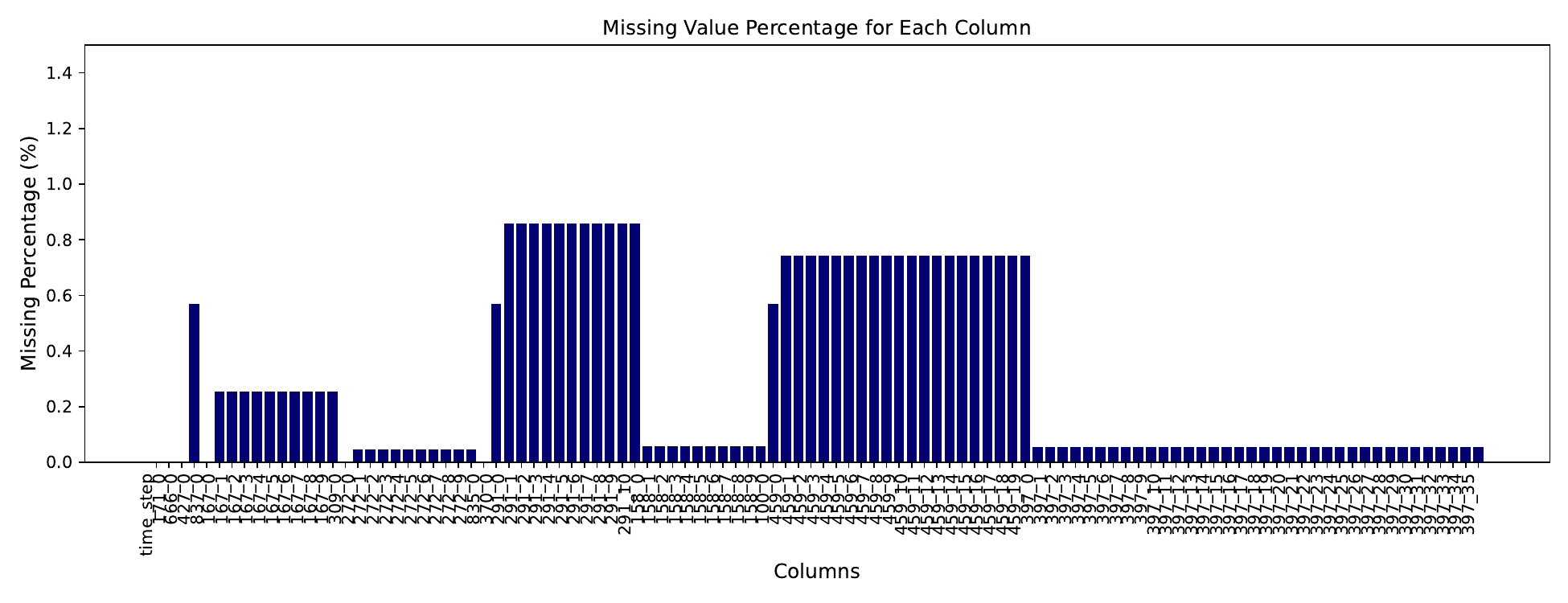}

\caption{The missing value percentage in train\_operational\_readouts.csv file is less than 1\% per feature column. Note that the y-axis shows the percentage of missing and is limited to 1.5 for better visualization.}
\label{fig:train missing}
\end{figure}

The distribution of missing values in \textit{train\_operational\_readouts.csv} is shown in Fig. \ref{fig:train missing}. 
In this dataset, all the readouts were collected directly from the ECUs connected to the sensors of the heavy-duty trucks. The missing values occurred for various reasons, such as the vehicle not being equipped with the sensor in question, the ECU software not being set to log the signal, or communication issues with the ECU. However, due to the high quality of the sensors and software, the rate of missing values is low, with less than 1 percent missingness per feature/column.
Given the low percentage of missing values, the dataset's integrity remains relatively intact, allowing various machine learning tasks with minimal preprocessing. 

\subsubsection{train\_tte.csv}
\begin{figure}[h!]
    \centering
    \includegraphics[width=\linewidth]{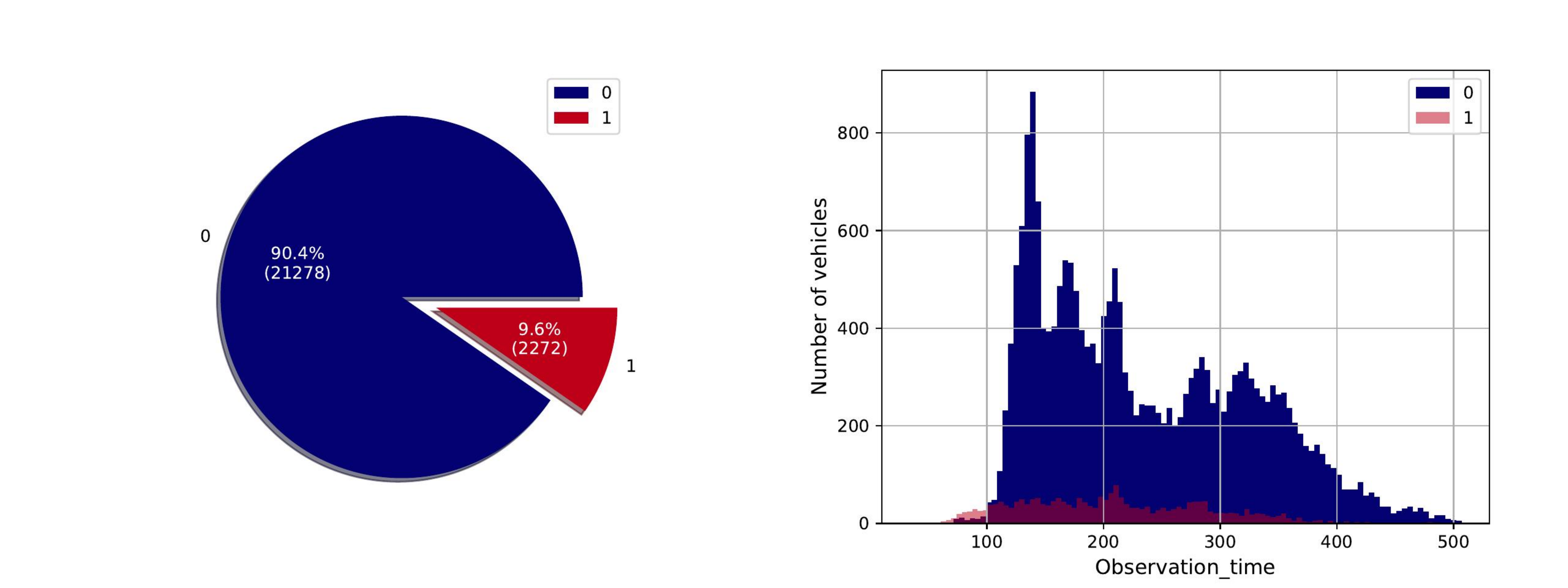}

    \begin{minipage}[t]{0.45\linewidth}
        \centering
        \vspace{-5mm}
        \subcaption{}\label{fig:4a}
    \end{minipage}
    \hfill
    \begin{minipage}[t]{0.45\linewidth}
        \centering
        \vspace{-5mm}
        \subcaption{}\label{fig:4b}
    \end{minipage}
    
    \caption{Visualization of classes in the train set. On the left, (a) shows the repair distribution, while on the right, (b) illustrates the distribution of healthy and failed components with different observation times during data collection.}
    \label{fig:combined-image4}
\end{figure}

The file with the name "\textit{train\_tte.csv}" contains the repair records of Component X collected from each vehicle, indicating the time\_to\_event (tte), i.e., the replacement time for Component X during the study period. This data file includes 23550 number of rows and two columns: "\textit{length\_of\_study\_time\_step}" and "\textit{in\_study\_repair}," where the former indicates the number of operation time steps after Component X started working. The latter is the class label, where it's set to 1 if Component X was repaired at the time equal to its corresponding \textit{length\_of\_study\_time\_step}, or it can take the value of zero in case no failure or repair event occurs during the first \textit{length\_of\_study\_time\_step} of operation.
It is good to mention that the "\textit{train\_tte.csv}" data is imbalanced with 21278 occurrences of label 0 and 2272 instances of label 1. In other words, it is skewed toward label 0. Figure \ref{fig:4a} compares the number of healthy and repaired components in the train set in general. Moreover, Fig. \ref{fig:4b} shows the distribution of healthy and repaired components in their corresponding observation time during the data collection, i.e., the time between the last and first readout for each vehicle.

\begin{figure}[ht]
\centering
\includegraphics[width=0.7\linewidth]{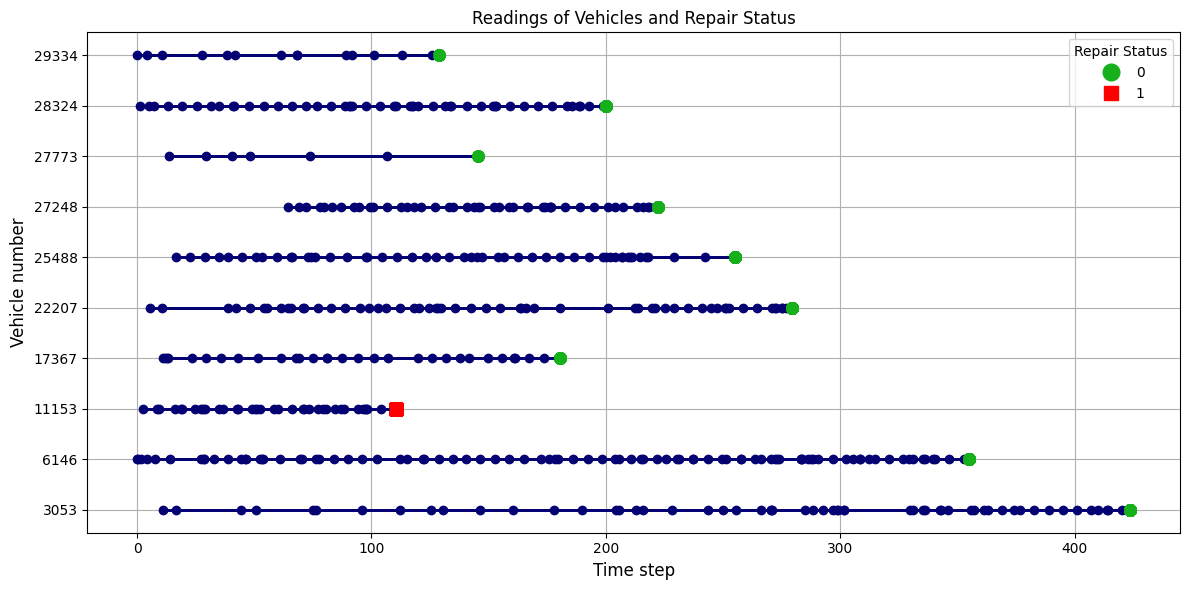}

\caption{History of readouts for ten random vehicles.}
\label{fig:tte}
\end{figure}

Delving deeper into the history of readouts in the training data, Fig. \ref{fig:tte} illustrates the time of the readouts for ten random vehicles. In this figure, blue dots represent the individual readout events for each vehicle and illustrate the monitoring frequency of events in the \textit{train\_operational\_readout} data file. Besides, the final readout for each vehicle is collected from the \textit{train\_tte} data part and highlighted with colors: Green circles show components having healthy status in their last workshop visits, while red squares mark repaired components. It also can be seen that the readouts are distributed unevenly among different vehicles.
In addition to this information, there are no missing values (shown by NaN) in this data file.

\subsubsection{train\_specifications.csv}

The last file in the training set is called "\textit{train\_specifications.csv}," which contains information about the specifications of the vehicles, such as their engine type and wheel configuration. In total, there are 23550 observations and eight categorical features, with no missing values for all vehicles.

\begin{figure}[h!]
\centering
\includegraphics[width=1\linewidth]{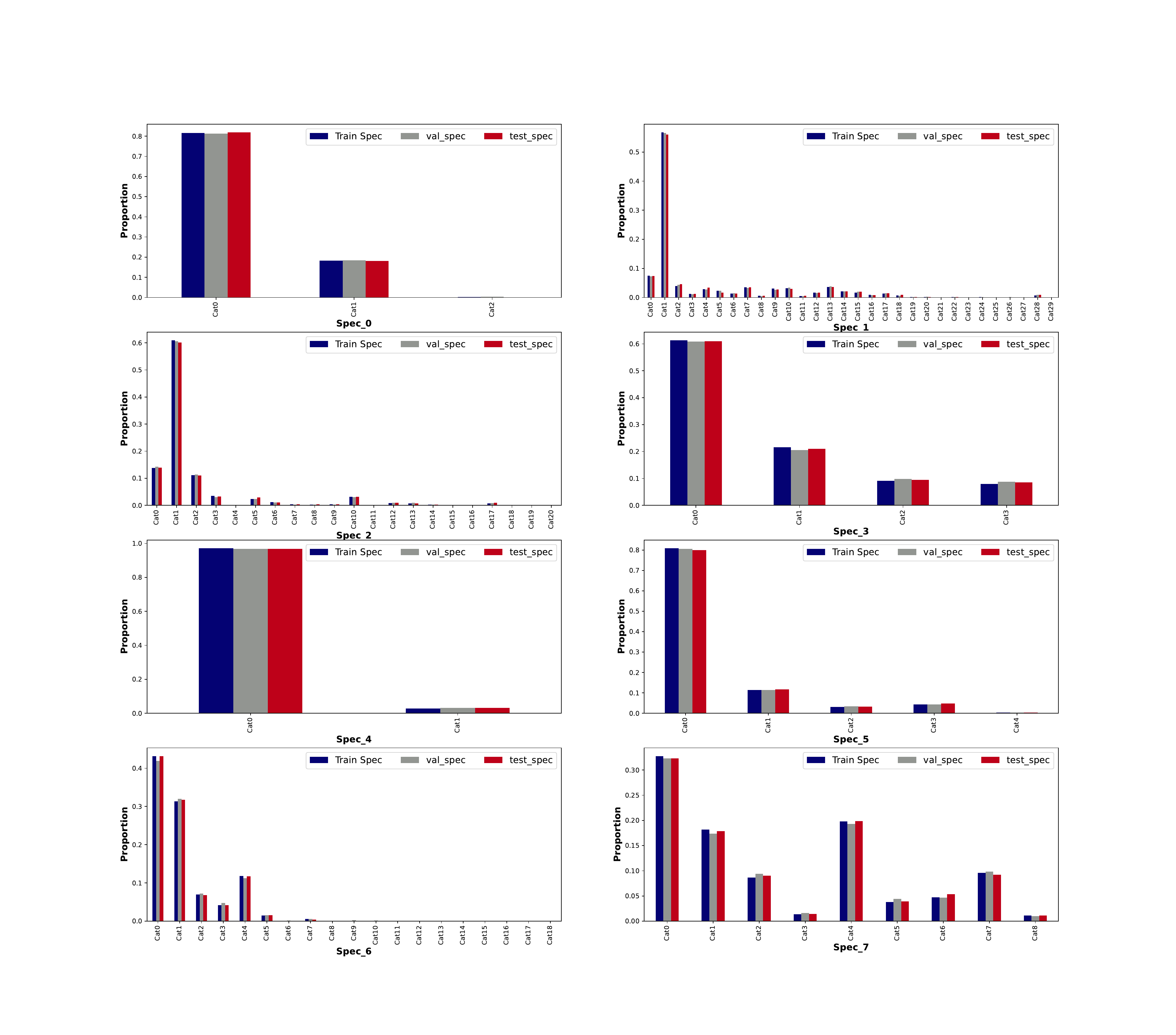}
\caption{Frequency analysis of specification (normalized) for training, validation, and test set.}
\label{fig:all spec norm}
\end{figure}

The features in \textit{train\_specifications.csv} are anonymized, each can take categories in {Cat0, Cat1, ..., Cat28}. Figure \ref{fig:all spec norm} illustrates the normalized frequency distribution of this categorical data for different data parts (train, validation, and test), allowing for comparison of how each category is represented across the three datasets. Each subplot represents a different specification (e.i., Spec\_0 to Spec\_7).
In each subplot, the x-axis shows the categorical values that each specification feature can take, and the y-axis depicts the normalized proportion of each category across the data parts with different colors: dark blue, grey, and red for train, validation, and test set, respectively. 
Since the proportions are similar across data parts, this indicates that the category distributions are relatively consistent.

\subsection{Validation set}

The validation set consists of three files called "\textit{validation\_labels.csv}", "\textit{validation\_operational\_readouts.csv}", 
and "\textit{validation\_specification.csv}".

\subsubsection{validation\_operational\_readouts.csv}\label{sec:val op}

\begin{figure}[h!]
\centering
\includegraphics[width=0.5\linewidth]{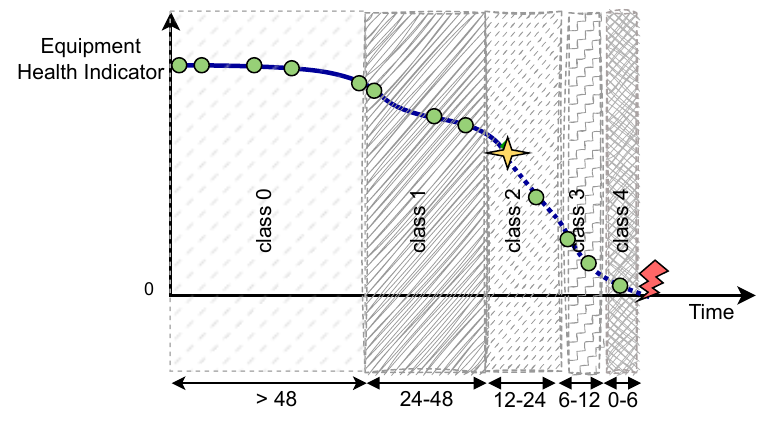}

\caption{Class interpretation in the validation and test data}
\label{fig:val class}
\end{figure}

In general, the \textit{validation\_operational\_readouts.csv} has the same description as the \textit{train\_operational\_readouts.csv} except for the fact that in \textit{validation\_operational\_readouts.csv}, 
the operational data is incomplete. Only a subset of the whole observations of each vehicle is provided, and it extends only up to a randomly selected readout. As a result, it lacks details about the entire lifespan of a vehicle. This is done to simulate the usage of a prediction model in a realistic scenario when it only has information about a vehicle up until the present time.
Figure \ref{fig:val class} illustrates an example of a hypothetical health indicator or degradation model of component X installed on a vehicle in the validation set. Green dots are recorded readouts from the start of its operation, and the yellow star represents the last simulated readout for this vehicle, which is randomly selected among all possible readouts. 
This means that we only have information up to that readout, and the rest of the information is not given.

Overall, it includes 196227 observations/rows showing the number of instances from 5046 vehicles and includes 107 columns. 
Moreover, similar to \textit{train\_operational\_readouts.csv}, it contains a minimal missing value (less than one percent for each feature).

\subsubsection{validation\_labels.csv}

The \textit{validation\_labels.csv} file has 5046 rows, which is equal to the number of vehicles contributed to the operational data of the validation set. It includes a column named \textit{class\_label}, corresponding to the class for the last readout of each vehicle. 
As mentioned in subsection \textit{validation\_operational\_readouts.csv}, the last readout for the validation set is selected randomly among all readouts for each vehicle. The temporal placement of this final simulated readout is categorized into five classes denoted by {0, 1, 2, 3, 4} where they are related to readouts within a time window of: (more than 48), (48 to 24), (24 to 12), (12 to 6), and (6 to 0) time\_step before the failure, respectively.
These classes show the time windows in which the last readouts for each vehicle are randomly selected. 
For instance, in Fig. \ref{fig:val class}, the last simulated readout is given in the time window of class 2.

This data set is also imbalanced and is skewed toward class 0, i.e., 4910 samples belong to class 0, while 76, 30, 16, and 14 samples belong to classes 4, 3, 1, and 2, respectively.

\begin{figure}[h!]
    \centering
    \includegraphics[width=\linewidth]{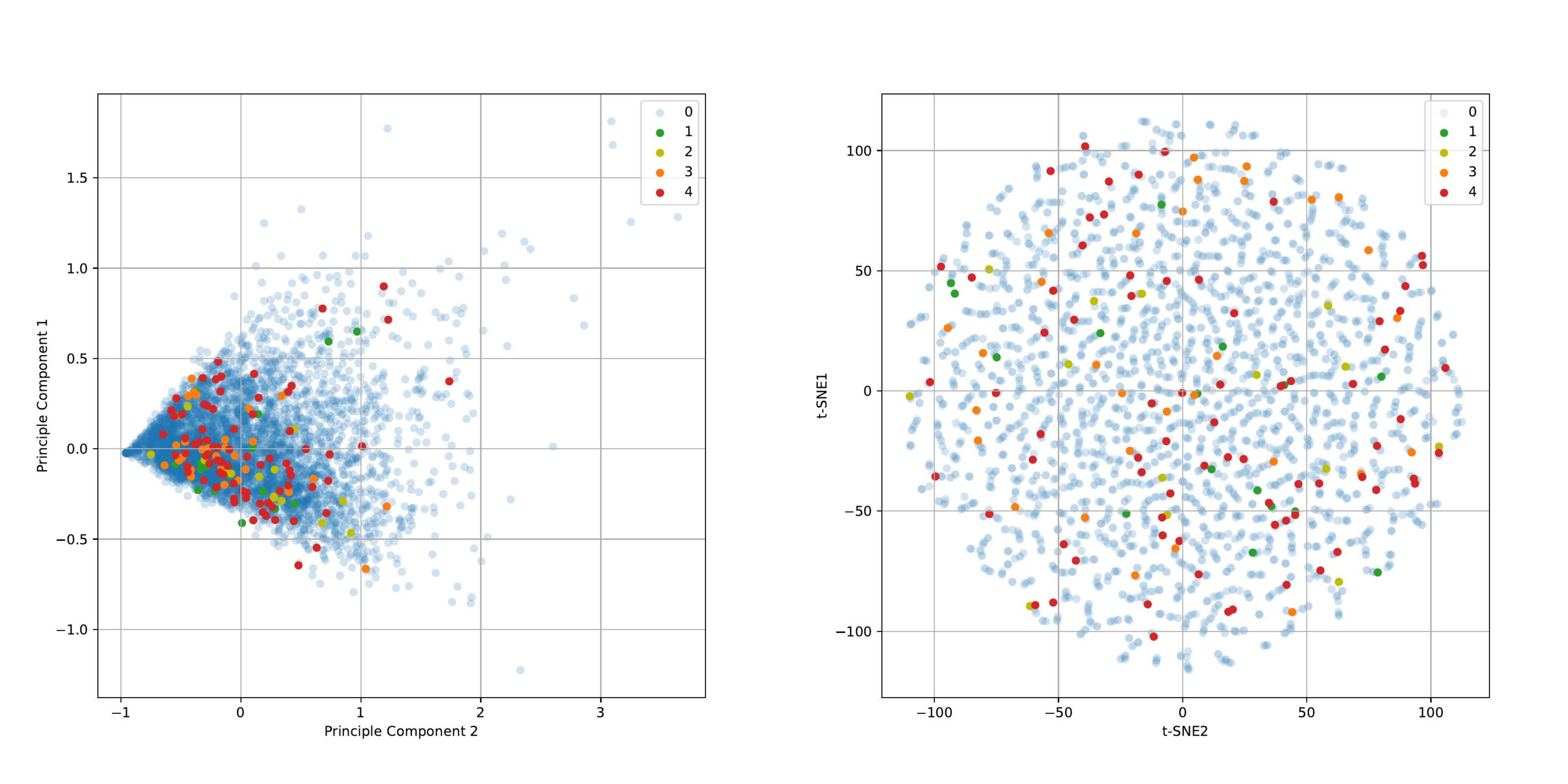}
    
    \begin{minipage}[t]{0.45\linewidth}
        \centering
        \vspace{-5mm}
        \subcaption{}\label{fig:8a}
    \end{minipage}
    \hfill
    \begin{minipage}[t]{0.45\linewidth}
        \centering
        \vspace{-5mm}
        \subcaption{}\label{fig:8b}
    \end{minipage}
    
    \caption{Visualization of validation data using dimensionality reduction techniques. Classes are shown in different colors. On the left, (a) presents the PCA analysis of the validation set, while on the right, (b) displays the t-SNE analysis.}
    \label{fig:combined-image8}
\end{figure}

For better visualization of the validation data, two techniques for dimensionality reduction, called Principal Component Analysis (PCA) and t-distributed Stochastic Neighbor Embedding (t-SNE), are performed on the last readout from each vehicle in the validation set. Figure \ref{fig:8a} and Fig. \ref{fig:8b} show the dataset when its dimension is reduced into two, using PCA and t-SNE techniques, respectively. Vehicles belonging to each class are shown in different colors. As can be seen, the five classes of the dataset are scrambled in the two-dimensional space both for PCA and t-SNE. This demonstrates the complexity of the problem when the features are projected into two dimensions.

\subsubsection{validation\_specification.csv}
The file \textit{validation\_specification.csv} has a similar structure as \textit{train\_specifications.csv} with no missing values, except the data is collected from 5046 vehicles in the validation set. Details of categorical values of each feature are shown in Fig. \ref{fig:all spec norm}. Compared to the bars related to the \textit{train\_specifications}, the grey bars related to the \textit{validation\_specification}, follows the same normalized distribution, where shows the consistency in categories across both data parts.

\subsection{Testing set}

The test set contains three files "\textit{test\_operational\_readouts.csv}", "\textit{test\_specifications.csv}", and "\textit{test\_labels.csv} which are explained in detail in the following sections."

\subsubsection{test\_operational\_readouts.csv}

Similar to \textit{validation\_operational\_readouts.csv}, for the \textit{test\_operational\_readouts.csv}, the last readouts of vehicles are randomly selected from the larger sequences observed during the study period (see Fig. \ref{fig:val class}) to emulates real-life situations. In summary, this file contains 198140 number of readouts from 14 variables (107 columns) gathered from 5045 unique vehicles.
Similar to \textit{validation\_operational\_readouts.csv} and \textit{train\_operational\_readouts.csv}, the percentage of missing values in this file is less than one percent for each feature.

\subsubsection{test\_labels.csv}
This data file includes the class label for 5045 vehicles in the test set that their last readouts are randomly selected in five classes of 0, 1, 2, 3, and 4. Like the \textit{validation\_labels}, this data file is also imbalanced and is skewed toward class 0, which contains 4903 samples. In comparison, classes 1, 2, 3, and 4 include 26, 15, 41, and 60 samples.

\subsubsection{test\_specifications.csv}

Specification information of 5045 test vehicles is collected in \textit{the test\_specifications.csv} file containing eight categorical features with values varying between Cat0, Cat1,..., and Cat28, with no missing values. 
Figure \ref{fig:all spec norm} illustrates the categories of each feature in the \textit{(*\_specifications.csv} files, including for the test set. Compared to the bars related to the train\_specifications and validation\_specifications, the red bars representing the test\_specification exhibit the same normalized distribution, indicating the consistency in category distribution across all three data parts.

\section{Technical Validation}

OEMs usually hesitate to share their device operational data for many reasons, such as GDPR compliance, confidentiality agreements (NDAs), ownership of data, device failure rates, etc. As a result, many works on data-driven PdM are limited to private datasets \cite{moat2021survival,rahat2023bridging,hoffmann2020roadmap,rahat2020modeling,bouabdallaoui2021predictive,rahat2022domain}. This prevents researchers who worked with real data comparing their methods performances with each other. The newly shared dataset \cite{Lindgren2024SCANIA} by SCANIA can serve as a benchmark in the field, making it easier for future researchers to compare their methods and generate reproducible results. 

The proposed dataset \cite{Lindgren2024SCANIA} is highly conducive to various machine learning tasks like regression, anomaly detection, survival analysis, and classification, in the PdM field. This merit arises from two supreme characteristics: being a real-world dataset collected from actual trucks and exhibiting a multi-variate time series structure.
In regression tasks, this dataset \cite{Lindgren2024SCANIA} is valuable for estimating components' remaining useful life or predicting the time until the next repair. Leveraging historical and time-to-event data, regression models can provide insightful estimates.
Survival analysis is a powerful technique that predicts the survival function and the probability of an event happening at a specific time while considering censored data in the training phase. The blue vehicle populations in Fig. \ref{fig:combined-image4} are referred to as censored data.
The temporal structure of the dataset is crucial for the effective application of survival analysis.
Furthermore, for classification tasks, using time series data, the model can classify whether a vehicle is going to fail within a specific time window or not. 

To evaluate the performance of the aforementioned models, a cost function is suggested by experts of the company, which is defined by the sum of the different “Cost$\_n\_m$” multiplied by the number of instances, resulting in a summarized cost (Total\_cost). 
\vspace{-0.2cm}
\begin{equation}\label{eq:1}
Total\_cost = Cost\_n\_m \times No\_instances
\end{equation}

Equation \ref{eq:1} calculates the total cost function, where $n$ shows the actual class, and m shows the predicted class, while $n, m$ $\in$ \{0,1,2,3,4\}. 
In general, when $n<m$, the Cost\_$n\_m$ indicates a cost for a false positive error, and if $n>m$, it indicates a cost for a false negative error. It should be emphasized that the cost of false negative prediction is much higher than that of false positive prediction.  Table \ref{tab:costs} demonstrates the different values of Cost$\_n\_m$ according to different values of actual and predicted class prediction. This table formulates a spectrum of the cost of failure from a catastrophic one (i.e., unexpected failure by the road) to a non-necessary cost of changing the component. The values of different costs in Table \ref{tab:costs} are defined by the experts of the company.
In this table, the term "Cost\_0\_\{1,2,3,4\}" denotes the cost incurred when a mechanic conducts an unnecessary check at a workshop. On the other hand, "Cost\_\{1,2,3,4\}$\_m$" represents the cost associated with the possibility of missing a faulty truck or triggering an alarm too late. This delay could lead to a breakdown or necessitate costly adjustments to the customer's transportation plan.

\begin{table}[ht]
\centering
\caption{Table of prediction cost}
\label{tab:costs}
\begin{tabular}{|l|l|l|l|l|l|}
\hline
 & Predicted: 0 & Predicted: 1 & Predicted: 2 & Predicted: 3 & Predicted: 4 \\ \hline
Actual: 0 &  & Cost\_0\_1=7 & Cost\_0\_2=8 & Cost\_0\_3=9 & Cost\_0\_4=10 \\ \hline
Actual: 1 & Cost\_1\_0=200 &  & Cost\_1\_2=7 & Cost\_1\_3=8 & Cost\_1\_4=9 \\ \hline
Actual: 2 & Cost\_2\_0=300 & Cost\_2\_1=200 &  & Cost\_2\_3=7 & Cost\_2\_4=8 \\ \hline
Actual: 3 & Cost\_3\_0=400 & Cost\_3\_1=300 & Cost\_3\_2=200 &  & Cost\_3\_4=7 \\ \hline
Actual: 4 & Cost\_4\_0=500 & Cost\_4\_1=400 & Cost\_4\_2=300 & Cost\_4\_3=200 &  \\ \hline
\end{tabular}
\end{table}

\section{Acknowledgements} 
This work has been funded by Scania CV AB and the Vinnova Program for Strategic Vehicle Research and Innovation (FFI).



\bibliographystyle{IEEEtran}  
\bibliography{references}  

\end{document}